# KINEMATICS OF A 3-*PRP* PLANAR PARALLEL ROBOT

Damien CHABLAT [*], Stefan STAICU [**]

*Articolul prezintă o modelare recurentă pentru cinematica unui robot paralel plan 3-PRP. Trei lanţuri cinematice plane ce conectează platforma mobilă a manipulatorului sunt situate în plan vertical. Cunoscând mişcarea platformei, se dezvoltă cinematica inversă şi se determină poziţiile, vitezele şi acceleraţiile robotului. Unele ecuaţii matriceale oferă expresi iterative şi grafice pentru deplasările, vitezele şi acceleraţiile celor trei acţionori de translaţie.*

*Recursive modelling for the kinematics of a 3-PRP planar parallel robot is presented in this paper. Three planar chains connecting to the moving platform of the manipulator are located in a vertical plane. Knowing the motion of the platform, we develop the inverse kinematics and determine the positions, velocities and accelerations of the robot. Several matrix equations offer iterative expressions and graphs for the displacements, velocities and accelerations of three prismatic actuators.*

## 1 Introduction

Compared with serial manipulators, potential advantages of the parallel architectures are higher kinematical precision, lighter weight, better stiffness, greater load bearing, stabile capacity and suitable positional actuator arrangements. However, they present limited workspace and complicated singularities [1].

Considerable efforts have been devoted to the kinematics and dynamic analysis of fully parallel manipulators. The class of manipulators known as Stewart-Gough platform focused great attention (Stewart [2]; Merlet [3]; Parenti-Castelli and Di Gregorio [4]) and is used in flight simulators and more recently for Parallel Kinematics Machines. The Delta parallel robot (Clavel [5]; Tsai and Stamper [6]; Staicu and Carp-Ciocardia [7]) as well as the Star parallel manipulator (Hervé and Sparacino [8]) are equipped with three motors, which train on the mobile platform in a three-degree-of-freedom general translation motion. Angeles, Gosselin, Gagné and Wang [9, 10, 11] analysed the kinematics, dynamics and singularity loci of a spherical robot with three actuators.

[*]Dr., Institut de Recherche en Communications et Cybernétique de Nantes, UMR CNRS 6597, FRANCE
[**]Prof., Department of Mechanics, University "Politehnica" of Bucharest, ROMANIA

A mechanism is considered a *planar robot* if all the moving links in the mechanism perform the planar motions; the loci of all points in all links can be drawn conveniently on a plane and the axes of the revolute joints must be normal to the plane of motion, while the direction of translation of a prismatic joint must be parallel to the plane of motion.

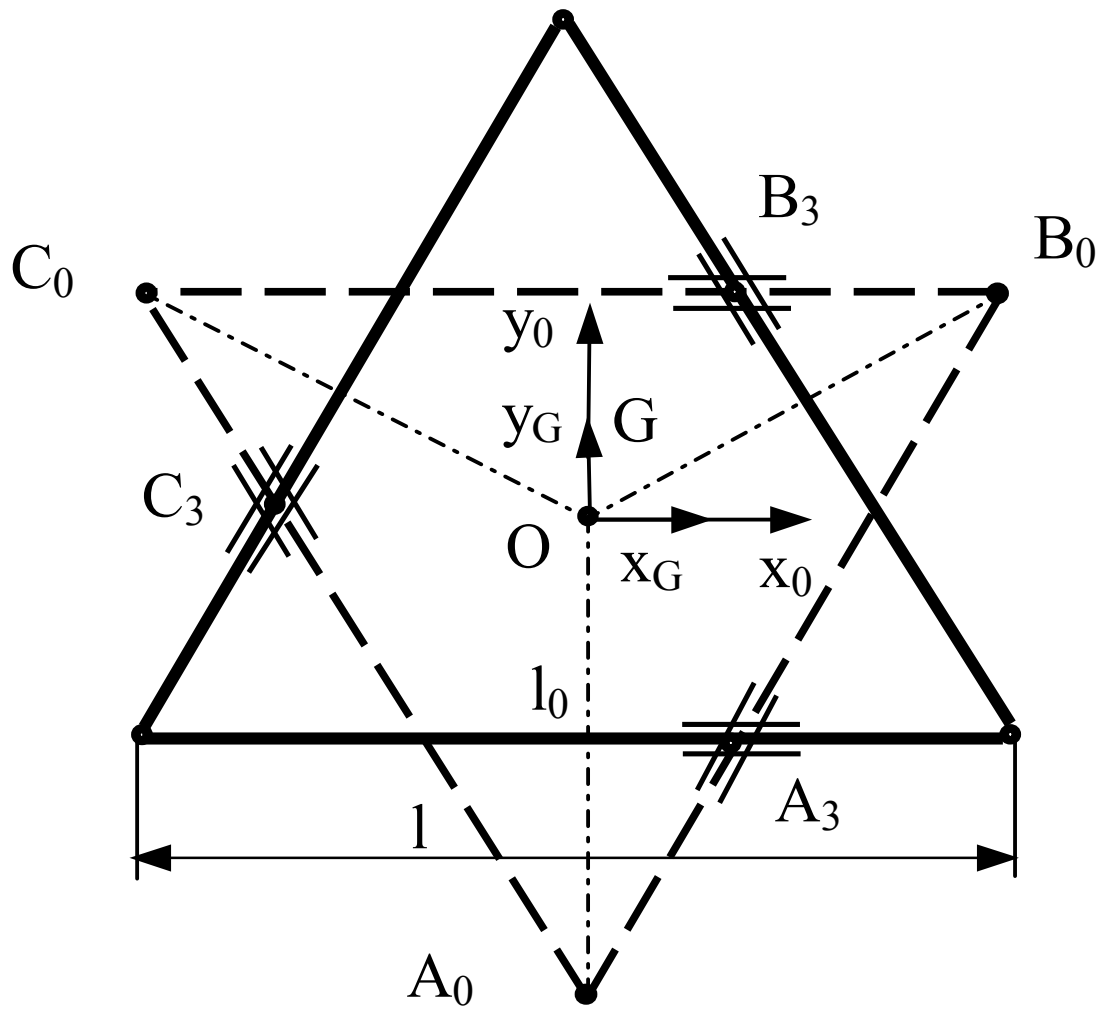

Fig. 1 The 3-*PRP* planar parallel robot

In their paper, Aradyfio and Qiao [12] examine the inverse kinematics solution for the three different 3-DOF planar parallel robots. Gosselin and Angeles [13] and Pennock and Kassner [14] each present a kinematical study of a planar parallel robot, where a moving platform is connected to a fixed base by three links, each leg consisting of two binary links and three parallel revolute joints. Sefrioui and Gosselin [15] give a numerical solution in the inverse and direct kinematics of this kind of robot. Mohammadi-Daniali et al. [16] present a study of velocity relationships and singular conditions for general planar parallel robots.

Merlet [17] solved the forward pose kinematics problem for a broad class of planar parallel manipulators. Williams and Reinholtz [18] analysed the dynamics and the control of a planar three-degree-of-freedom parallel manipulator at Ohio University, while Yang et al. [19] concentrate on the singularity analysis of a class of 3-*RRR* planar parallel robots developed in their laboratory. Bonev, Zlatanov and Gosselin [20] describe several types of singular configurations by studying the direct kinematics model of a 3-*RPR* planar parallel robot with actuated base joints. Mohammadi-Daniali et al. [21] analysed the kinematics of a planar 3-DOF parallel manipulator using the three *PRP* legs, where the three revolute joint axes are perpendicular to the plane of motion while the prismatic joint axes lie on the same plane.

A recursive method is developed in the present paper for deriving the inverse kinematics of the 3-*PRP* planar parallel robot in a numerically efficient way.

## 2 Kinematics modelling

The planar 3-*PRP* parallel robot is a special symmetrical closed-loop mechanism composed of three planar kinematical chains with identical topology, all connecting the fixed base to the moving platform. Three points $A_0, B_0, C_0$ represent the summits of a fixed triangular base and other three points define the geometry of the moving platform. Each leg consists of two links, with one revolute and two prismatic joints. The parallel mechanism with seven links $(T_k, k = 1, 2, ..., 7)$ consists of three revolute and six prismatic joints (Fig.1). Grübler mobility equation predicts that the device has certainly three degrees of freedom.

In the actuation scheme *PRP* each prismatic joint is an actively controlled prismatic cylinder. Thus, all prismatic actuators can be located on the fixed base. We attach a Cartesian frame $x_0 y_0 z_0 (T_0)$ to the fixed base with its origin located at triangle centre $O$, the $z_0$ axis perpendicular to the base and the $x_0$ axis pointing along the $C_0 B_0$ direction. Another mobile reference frame $x_G y_G z_G$ is attached to the moving platform. The origin of the central reference frame $x_G y_G z_G$ is located at the centre $G$ of the moving triangle (Fig. 2).

In the following we will represent the intermediate reference systems by only two axes so as is used in many robotics papers [1, 3, 9] and we note that the relative translation $\lambda_{k,k-1}$ and the rotation angle $\varphi_{k,k-1}$ point along or about the direction of $z_k$ axis.

We consider that the moving platform is initially located at a *central configuration*, where the platform is not rotated with respect to the fixed base and the mass centre $G$ is at the origin $O$ of fixed frame. One of three active legs (for example leg $A$) consists of a prismatic joint, which is as well as a linear drive **1** linked at the $x_1^A y_1^A z_1^A$ frame, having a rectilinear motion of displacement $\lambda_{10}^A$, velocity $v_{10}^A = \dot{\lambda}_{10}^A$ and acceleration $\gamma_{10}^A = \ddot{\lambda}_{10}^A$. Following link of the leg is a rigid body **2** linked at the $x_2^A y_2^A z_2^A$ frame, having a relative rotation about $z_2^A$ axis with the angle $\varphi_{21}^A$, velocity $\omega_{21}^A = \dot{\varphi}_{21}^A$ and acceleration $\varepsilon_{21}^A = \ddot{\varphi}_{21}^A$. A prismatic passive joint is introduced at a planar moving platform as an equilateral triangle with the edge $l = l_0\sqrt{3}$, which relatively translates at the displacement $\lambda_{32}^A$ and the velocity $v_{32}^A = \dot{\lambda}_{32}^A$ along $z_3^A$ axis.

Also, we consider that at the central configuration all legs are symmetrically extended and that the angles of orientation of three edges of fixed platform are given by

$$\alpha_A = \frac{\pi}{3}, \alpha_B = \pi, \alpha_C = -\frac{\pi}{3}. \tag{1}$$

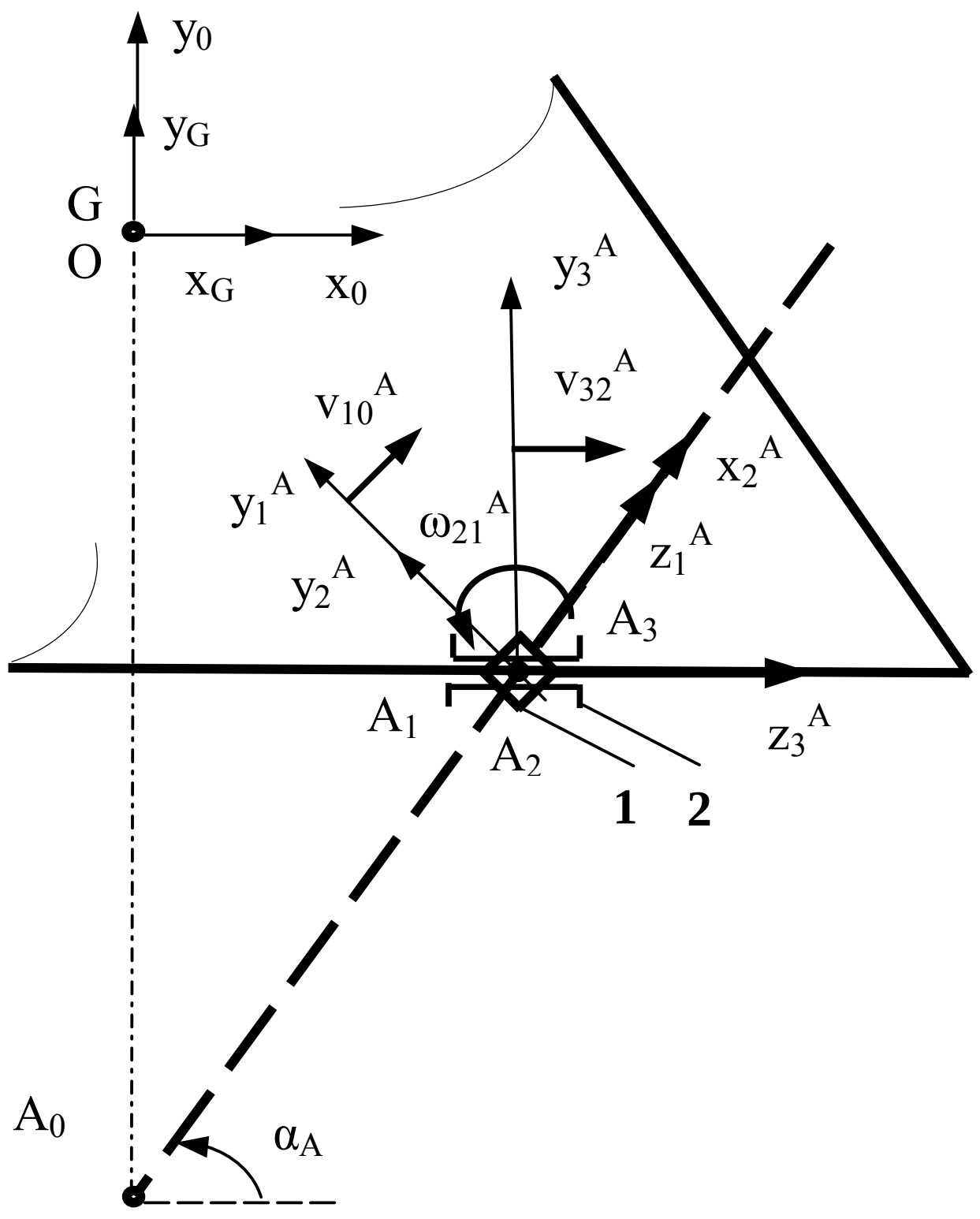

Fig. 2 Kinematical scheme of first leg $A$ of the mechanism

In the following, we apply the method of successive displacements to geometric analysis of closed-loop chains and we note that a joint variable is the displacement required to move a link from the initial location to the actual position. If every link is connected to least two other links, the chain forms one or more independent closed-loops.

The variable angle $\varphi^i_{k,k-1}$ of rotation about the joint axis $z^i_k$ is the parameter needed to bring the next link from a reference configuration to the next configuration. We call the matrix $q^{\varphi}_{k,k-1}$, for example, the orthogonal transformation $3\times3$ matrix of relative rotation with the angle $\varphi^i_{k,k-1}$ of link $T^i_k$ around $z^i_k$.

In the study of the kinematics of robot manipulators, we are interested in deriving a matrix equation relating the location of an arbitrary $T_k^i$ body to the joint variables. When the change of coordinates is successively considered, the corresponding matrices are multiplied. So, starting from the reference origin $O$ and pursuing the three legs $OA_0A_1A_2A_3$, $OB_0B_1B_2B_3$, $OC_0C_1C_2C_3$, we obtain the following transformation matrices [22]:

$$q_{10} = \theta_1 \theta_\alpha^i, \; q_{21} = q_{21}^\varphi \theta_1^T, \; q_{32} = \theta_1 \theta_2 \quad \text{with } (q = a,\ b,\ c),\ (i = A,\ B,\ C) \tag{2}$$

where $$q_{k,k-1}^\varphi = \begin{bmatrix} \cos\varphi_{k,k-1}^i & \sin\varphi_{k,k-1}^i & 0 \\ -\sin\varphi_{k,k-1}^i & \cos\varphi_{k,k-1}^i & 0 \\ 0 & 0 & 1 \end{bmatrix}, \; \theta_\alpha^i = \begin{bmatrix} \cos\alpha_i & \sin\alpha_i & 0 \\ -\sin\alpha_i & \cos\alpha_i & 0 \\ 0 & 0 & 1 \end{bmatrix}$$

$$\theta_1 = \begin{bmatrix} 0 & 0 & -1 \\ 0 & 1 & 0 \\ 1 & 0 & 0 \end{bmatrix}, \; \theta_2 = \frac{1}{2}\begin{bmatrix} 1 & -\sqrt{3} & 0 \\ \sqrt{3} & 1 & 0 \\ 0 & 0 & 2 \end{bmatrix} \tag{3}$$

$$q_{k0} = \prod_{s=1}^{k} q_{k-s+1,k-s} \quad (k = 1, 2, 3).$$

The displacements $\lambda_{10}^A$, $\lambda_{10}^B$, $\lambda_{10}^C$ of the active links are the joint variables that give the input vector $\vec{\lambda}_{10} = [\lambda_{10}^A \;\; \lambda_{10}^B \;\; \lambda_{10}^C]^T$ of the position of the mechanism. But, in the inverse geometric problem, we can consider that the position of the mechanism is completely given by the coordinates $x_0^G$, $y_0^G$ of the mass centre $G$ of the moving platform and the orientation angle $\phi$ of the movable frame $x_G y_G z_G$. The orthogonal rotation matrix of the moving platform from $x_0 y_0 z_0$ to $x_G y_G z_G$ reference system is

$$R = \begin{bmatrix} \cos\phi & \sin\phi & 0 \\ -\sin\phi & \cos\phi & 0 \\ 0 & 0 & 1 \end{bmatrix}. \tag{4}$$

Further, we suppose that the position vector $\vec{r}_0^{\,G} = [x_0^G \;\; y_0^G \;\; 0]^T$ of $G$ centre and the $\phi$ orientation angle, which are expressed by following analytical functions

$$\frac{x_0^G}{x_0^{G*}} = \frac{y_0^G}{y_0^{G*}} = \frac{\phi}{\phi^*} = 1 - \cos\frac{\pi}{3}t\,, \tag{5}$$

can describe the general absolute motion of the moving platform in its *vertical plane*. The values $2x_0^{G*}, 2y_0^{G*}, 2\phi^*$ denote the final position of the moving platform.

The conditions concerning the absolute orientation of the moving platform are expressed by three identities

$$\overset{\circ}{q}{}_{30}^{T} q_{30} = R, \quad (q = a, b, c) \tag{6}$$

where the resulting matrix $q_{30}$ is obtained by multiplying three basic matrices

$$q_{30} = q_{32}q_{21}q_{10}\,,\ \overset{\circ}{q}_{30} = q_{30}(t=0) = \theta_1\theta_2\theta_\alpha^i \quad (i = A, B, C)\,. \tag{7}$$

From these conditions one obtains the first relations between the angles of rotation

$$\varphi_{21}^A = \varphi_{21}^B = \varphi_{21}^C = \phi\,. \tag{8}$$

Six independent variables $\lambda_{10}^A, \lambda_{32}^A, \lambda_{10}^B, \lambda_{32}^B, \lambda_{10}^C, \lambda_{32}^C$ will be determined by several vector-loop equations as follows

$$\vec{r}_{10}^{\,i} + \sum_{k=1}^{2} q_{k0}^T\, \vec{r}_{k+1,k}^{\,i} + q_{30}^T\, \vec{r}_3^{\,Gi} = \vec{r}_0^{\,G}\,,\ (q = a, b, c)\ (i = A, B, C) \tag{9}$$

where

$$\vec{r}_{10}^{\,i} = \vec{r}_{00}^{\,i} + (l_0/\sqrt{3} + \lambda_{10}^i)q_{10}^T\vec{u}_3\,,\ \vec{r}_{00}^{\,A} = l_0[0 \quad -1 \quad 0]^T$$

$$\vec{r}_{00}^{\,B} = \frac{1}{2}l_0[\sqrt{3} \ \ 1 \ \ 0]^T,\ \vec{r}_{00}^{\,C} = \frac{1}{2}l_0[-\sqrt{3} \ \ 1 \ \ 0]^T$$

$$\vec{r}_{21}^{\,i} = \vec{0},\ \vec{r}_{32}^{\,i} = \lambda_{32}^i q_{32}^T\vec{u}_3\,,\ \vec{r}_3^{\,Gi} = \frac{1}{2}l_0[0 \ \ 1 \ \ -\frac{\sqrt{3}}{3}]^T$$

$$\vec{u}_1 = \begin{bmatrix}1\\0\\0\end{bmatrix}, \vec{u}_2 = \begin{bmatrix}0\\1\\0\end{bmatrix}, \vec{u}_3 = \begin{bmatrix}0\\0\\1\end{bmatrix}, \tilde{u}_3 = \begin{bmatrix}0 & -1 & 0\\1 & 0 & 0\\0 & 0 & 0\end{bmatrix}. \tag{10}$$

Actually, these vector equations mean that

$$\left(\frac{l_0}{\sqrt{3}} + \lambda_{10}^i\right)\cos\alpha_i + \lambda_{32}^i\cos\left(\phi - \frac{\pi}{3} + \alpha_i\right) = x_0^G - x_{00}^i - \frac{l_0}{2\sqrt{3}}\cos(\phi + \alpha_i) + \frac{l_0}{2}\sin(\phi + \alpha_i)$$

$$\left(\frac{l_0}{\sqrt{3}}+\lambda_{10}^i\right)\sin\alpha_i+\lambda_{32}^i\sin\left(\phi-\frac{\pi}{3}+\alpha_i\right)=y_0^G-y_{00}^i-\frac{l_0}{2\sqrt{3}}\sin(\phi+\alpha_i)-\frac{l_0}{2}\cos(\phi+\alpha_i) \quad (11)$$

with $(i=A,\ B,\ C)$.

Developing the inverse kinematics problem, we determine the velocities and accelerations of the manipulator, supposing that the planar motion of the moving platform is known. So, we compute the linear and angular velocities of each leg in terms of the angular velocity $\vec{\omega}_0^G=\dot{\phi}\vec{u}_3$ and the centre's velocity $\vec{v}_0^G=\dot{\vec{r}}_0^G$ of the moving platform.

The rotation motions of the elements of each leg (leg $A$, for example) are characterized by recursive relations of following skew-symmetric matrices

$$\tilde{\omega}_{k0}^A=a_{k,k-1}\tilde{\omega}_{k-1,0}^A a_{k,k-1}^T+\omega_{k,k-1}^A\tilde{u}_3,\ \omega_{k,k-1}^A=\dot{\varphi}_{k,k-1}^A\quad (k=1,2,3), \quad (12)$$

which are *associated* to the absolute angular velocities

$$\vec{\omega}_{10}^A=\vec{0},\vec{\omega}_{20}^A=a_{21}\vec{\omega}_{10}^A+\vec{\omega}_{21}^A=\dot{\phi}\vec{u}_3,\ \vec{\omega}_{30}^A=a_{32}\vec{\omega}_{20}^A+\vec{\omega}_{32}^A=\dot{\phi}\vec{u}_3. \quad (13)$$

Following relations give the velocities $\vec{v}_{k0}^A$ of joints $A_k$

$$\vec{v}_{10}^A=\dot{\lambda}_{10}^A\vec{u}_3,\ \vec{v}_{21}^A=\vec{0},\ \vec{v}_{32}^A=\dot{\lambda}_{32}^A\vec{u}_3$$

$$\vec{v}_{k0}^A=a_{k,k-1}\vec{v}_{k-1,0}^A+a_{k,k-1}\tilde{\omega}_{k-1,0}^A\vec{r}_{k,k-1}^A+v_{k,k-1}^A\vec{u}_3. \quad (14)$$

The geometrical equations of constraints (8) and (9) when differentiated with respect to time lead to the following *matrix conditions of connectivity* [23]

$$v_{10}^A\vec{u}_j^T a_{10}^T\vec{u}_3+v_{32}^A\vec{u}_j^T a_{30}^T\vec{u}_3=\vec{u}_j^T\dot{\vec{r}}_0^G-\omega_{21}^A\vec{u}_j^T\{\lambda_{32}^A a_{20}^T\tilde{u}_3 a_{32}^T\vec{u}_3+a_{20}^T\tilde{u}_3 a_{32}^T\vec{r}_3^{GA}\}\ \ (j=1,2)$$

$$\omega_{21}^A=\dot{\phi}, \quad (15)$$

where $\tilde{u}_3$ is a skew-symmetric matrix associated to unit vector $\vec{u}_3$ pointing in the positive direction of $z_k$ axis. From these equations, we obtain the relative velocities $v_{10}^A, \omega_{21}^A, v_{32}^A$ as functions of angular velocity of the platform and velocity of mass centre $G$ and the *complete* Jacobian matrix of the manipulator. This matrix is a fundamental element for the analysis of the robot workspace and the particular configurations of singularities where the manipulator becomes uncontrollable.

By rearranging, the derivatives with respect to time of the six constraint equations (11) lead to the matrix equation

$$J_1\dot{\lambda}_{10}=J_2[\dot{x}_0^G\quad \dot{y}_0^G\quad \dot{\phi}]^T \quad (16)$$

for the planar robot with prismatic actuators.

The matrices $J_1$ and $J_2$ are the inverse and forward Jacobian of the manipulator and can be expressed as

$$J_1 = diag\,\{\delta_A \quad \delta_B \quad \delta_C\}$$

$$J_2 = \begin{bmatrix} \beta_1^A & \beta_2^A & \beta_3^A \\ \beta_1^B & \beta_2^B & \beta_3^B \\ \beta_1^C & \beta_2^C & \beta_3^C \end{bmatrix}, \tag{17}$$

with

$$\delta_i = \sin(\phi - \frac{\pi}{3}) \quad (i = A, B, C)$$

$$\beta_1^i = \sin(\phi - \frac{\pi}{3} + \alpha_i), \beta_2^i = -\cos(\phi - \frac{\pi}{3} + \alpha_i) \tag{18}$$

$$\begin{aligned} \beta_3^i = {} & (x_0^G - x_{00}^i)\cos(\phi - \frac{\pi}{3} + \alpha_i) + \\ & + (y_0^G - y_{00}^i)\sin(\phi - \frac{\pi}{3} + \alpha_i) - \\ & - (\frac{l_0}{\sqrt{3}} + \lambda_{10}^i)\cos(\phi - \frac{\pi}{3}). \end{aligned}$$

The singular configurations of the three closed-loop kinematical chains can easily be determined through the analysis of two Jacobian matrices $J_1$ and $J_2$ [24, 25]. For the matrix $J_1$, the determinant vanishes if $\phi = 2\pi/3$, which leads also to a singular configuration of $J_2$.

Concerning the relative accelerations $\gamma_{10}^A, \varepsilon_{21}^A, \gamma_{32}^A$ of the robot, new connectivity conditions are obtained by the time derivative of equations in (15), which are [26]

$$\begin{aligned} & \gamma_{10}^A \vec{u}_j^T a_{10}^T \vec{u}_3 + \gamma_{32}^A \vec{u}_j^T a_{30}^T \vec{u}_3 = \vec{u}_j^T \ddot{\vec{r}}_0^G - \\ & - \omega_{21}^A \omega_{21}^A \vec{u}_j^T \{a_{20}^T \tilde{u}_3 \tilde{u}_3 a_{32}^T \vec{u}_3 + a_{20}^T \vec{u}_3 \tilde{u}_3 a_{32}^T \vec{r}_3^{GA}\} - \\ & - \varepsilon_{21}^A \vec{u}_j^T \{\lambda_{32}^A a_{20}^T \tilde{u}_3 a_{32}^T \vec{u}_3 + a_{20}^T \tilde{u}_3 a_{32}^T \vec{r}_3^{GA}\} - 2\omega_{21}^A v_{32}^A \vec{u}_j^T a_{20}^T \tilde{u}_3 a_{32}^T \vec{u}_3 \quad (j = 1, 2) \end{aligned}$$

$$\varepsilon_{21}^A = \dot{\phi}\,. \tag{19}$$

The formulations in (15) and (19) are for $A$ only and they also apply to the legs $B$ and $C$, if the superscript $A$ is replaced by either $B$ or $C$.

Following recursive relations give the angular accelerations $\vec{\varepsilon}_{k0}^{A}$ and the accelerations $\vec{\gamma}_{k0}^{A}$ of joints $A_k$:

$$\vec{\gamma}_{10}^{A} = \ddot{\lambda}_{10}^{A}\vec{u}_3,\ \vec{\gamma}_{21}^{A} = \vec{0},\ \vec{\gamma}_{32}^{A} = \ddot{\lambda}_{32}^{A}\vec{u}_3$$

$$\vec{\varepsilon}_{10}^{A} = \vec{0},\ \vec{\varepsilon}_{21}^{A} = \ddot{\phi}\vec{u}_3,\ \vec{\varepsilon}_{32}^{A} = \vec{0}$$

$$\vec{\varepsilon}_{k0}^{A} = a_{k,k-1}\vec{\varepsilon}_{k-1,0}^{A} + \varepsilon_{k,k-1}^{A}\vec{u}_3 + \omega_{k,k-1}^{A}a_{k,k-1}\tilde{\omega}_{k-1,0}^{A}a_{k,k-1}^{T}\vec{u}_3$$

$$\begin{aligned}\tilde{\omega}_{k0}^{A}\tilde{\omega}_{k0}^{A} + \tilde{\varepsilon}_{k0}^{A} &= a_{k,k-1}\left(\tilde{\omega}_{k-1,0}^{A}\tilde{\omega}_{k-1,0}^{A} + \tilde{\varepsilon}_{k-1,0}^{A}\right)a_{k,k-1}^{T} + \omega_{k,k-1}^{A}\omega_{k,k-1}^{A}\tilde{u}_3\tilde{u}_3 + \varepsilon_{k,k-1}^{A}\tilde{u}_3 + \\ &+2\omega_{k,k-1}^{A}a_{k,k-1}\tilde{\omega}_{k-1,0}^{A}a_{k,k-1}^{T}\tilde{u}_3\end{aligned} \tag{20}$$

$$\begin{aligned}\vec{\gamma}_{k0}^{A} &= a_{k,k-1}\vec{\gamma}_{k-1,0}^{A} + a_{k,k-1}(\tilde{\omega}_{k-1,0}^{A}\tilde{\omega}_{k-1,0}^{A} + \tilde{\varepsilon}_{k-1,0}^{A})\vec{r}_{k,k-1}^{A} + \\ &+ 2v_{k,k-1}^{A}a_{k,k-1}\tilde{\omega}_{k-1,0}^{A}a_{k,k-1}^{T}\vec{u}_3 + \vec{\gamma}_{k,k-1}^{A}\vec{u}_3,\ (k = 1, 2, 3)\end{aligned}$$

We can notice that for this robot, the displacement of the leg is very simple. The displacement of body 2 is only along a fixed line, and its velocity and acceleration is equal to those of actuated joint associated.

For simulation purposes let us consider a planar robot, which has the following characteristics:

$$x_0^{G*} = 0.025\,m,\ y_0^{G*} = 0.025\,m,\ \phi^* = \frac{\pi}{12},\ \Delta t = 3\ s$$

$$l_0 = OA_0 = OB_0 = OC_0 = 0.3m,\ l = l_0\sqrt{3}$$

A program, which implements the suggested algorithm, is developed in MATLAB to solve the inverse kinematics of the planar PRP parallel robot. For illustration, it is assumed that for a period of three second the platform starts at rest from a central configuration and rotates or moves along two orthogonal directions. A numerical study of the robot kinematics is carried out by computation of the displacements $\lambda_{10}^{A},\ \lambda_{10}^{B},\ \lambda_{10}^{C}$, the velocities $v_{10}^{A},\ v_{10}^{B},\ v_{10}^{C}$, and the accelerations $\gamma_{10}^{A},\ \gamma_{10}^{B},\ \gamma_{10}^{C}$ of three prismatic actuators.

Following examples are solved to illustrate the simulation. For the first example we consider the *rotation motion* of the moving platform about $z_0$ axis with variable angular acceleration while all the other positional parameters are held equal to zero.

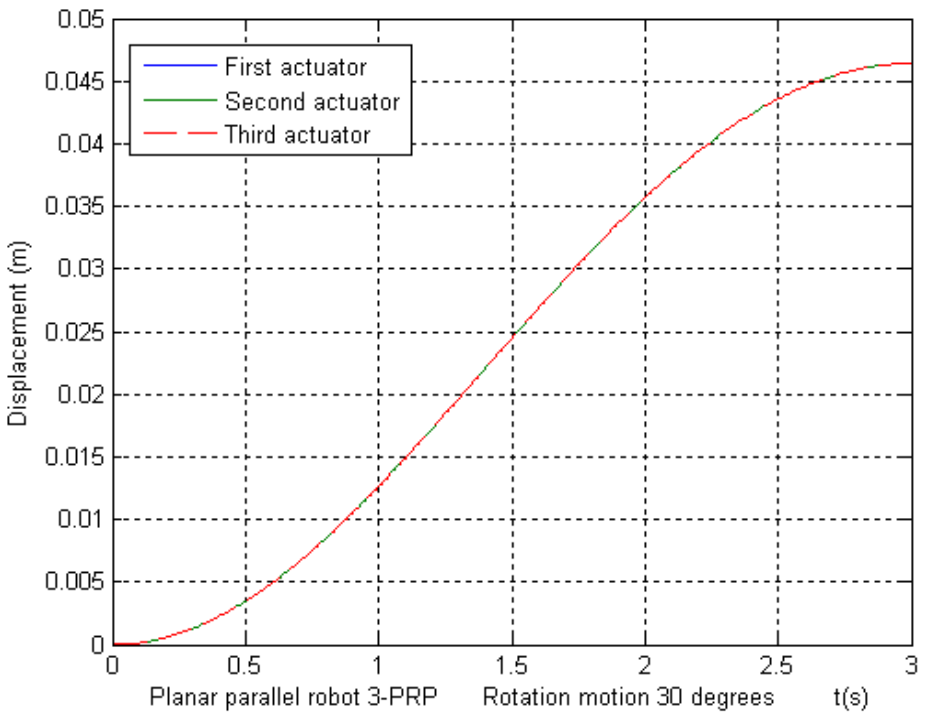

Fig. 3 Displacements $\lambda_{10}^{A}$, $\lambda_{10}^{B}$, $\lambda_{10}^{C}$

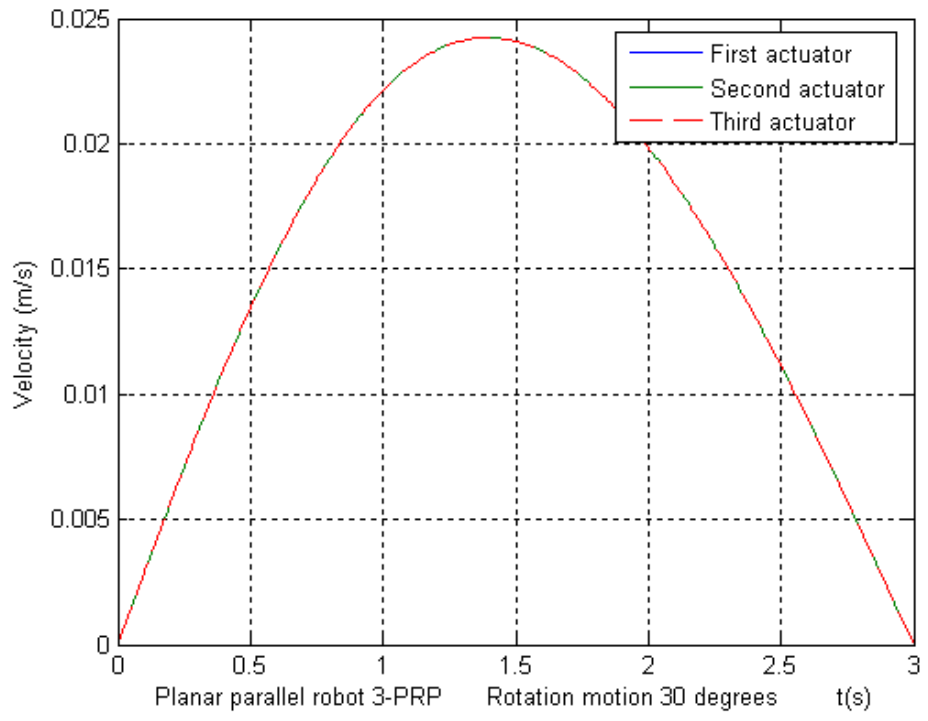

Fig. 4 Velocities $v_{10}^{A}$, $v_{10}^{B}$, $v_{10}^{C}$

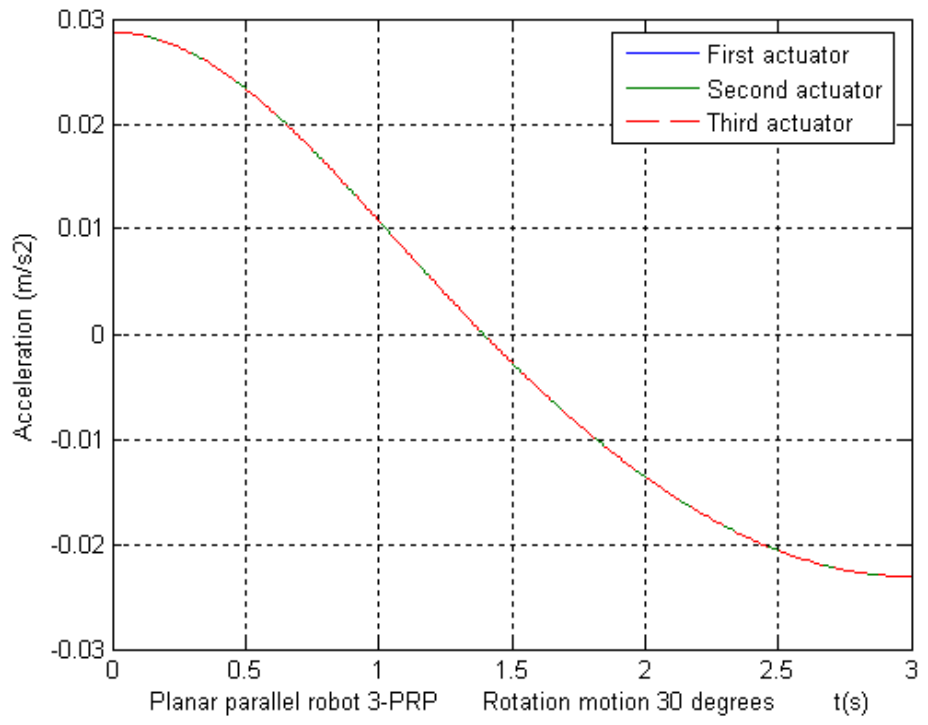

Fig. 5 Accelerations $\gamma_{10}^{A}$, $\gamma_{10}^{B}$, $\gamma_{10}^{C}$

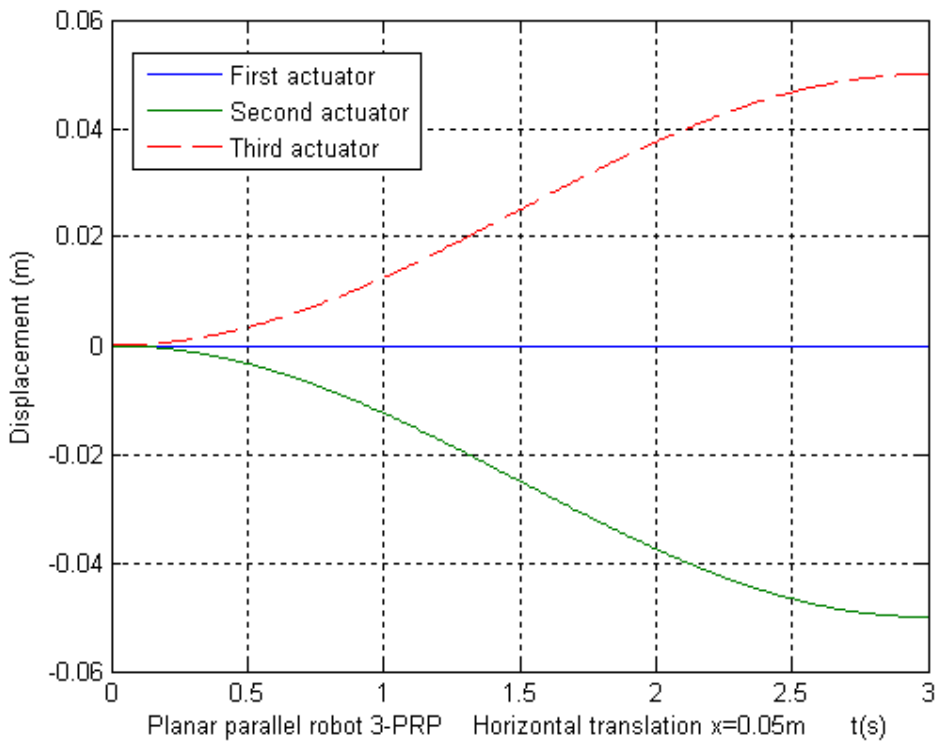

Fig. 6 Displacements $\lambda_{10}^{A}$, $\lambda_{10}^{B}$, $\lambda_{10}^{C}$

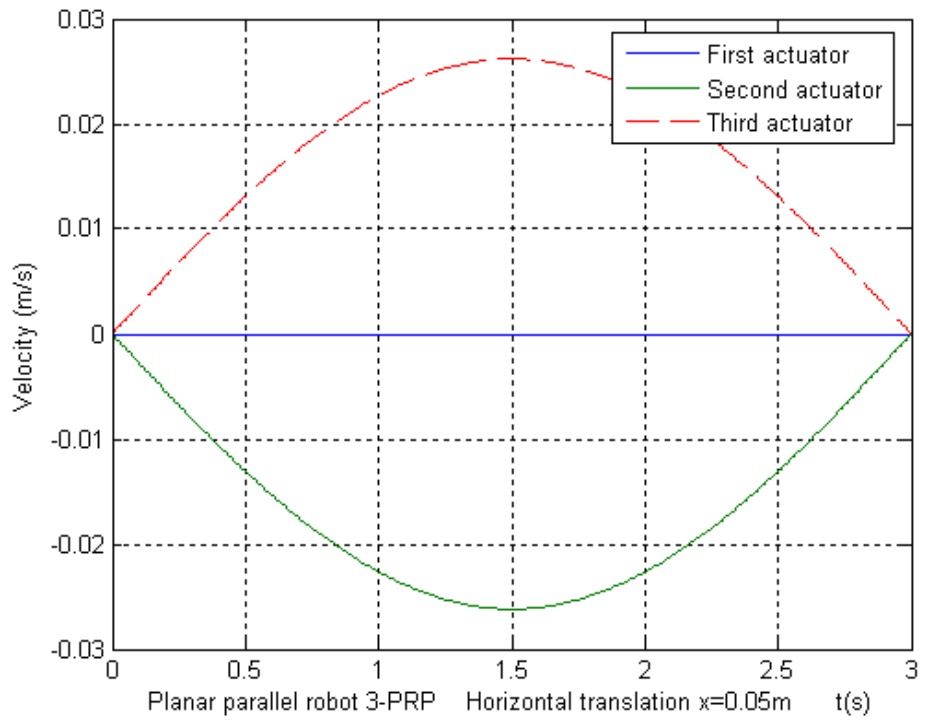

Fig. 7 Velocities $v_{10}^{A}$, $v_{10}^{B}$, $v_{10}^{C}$

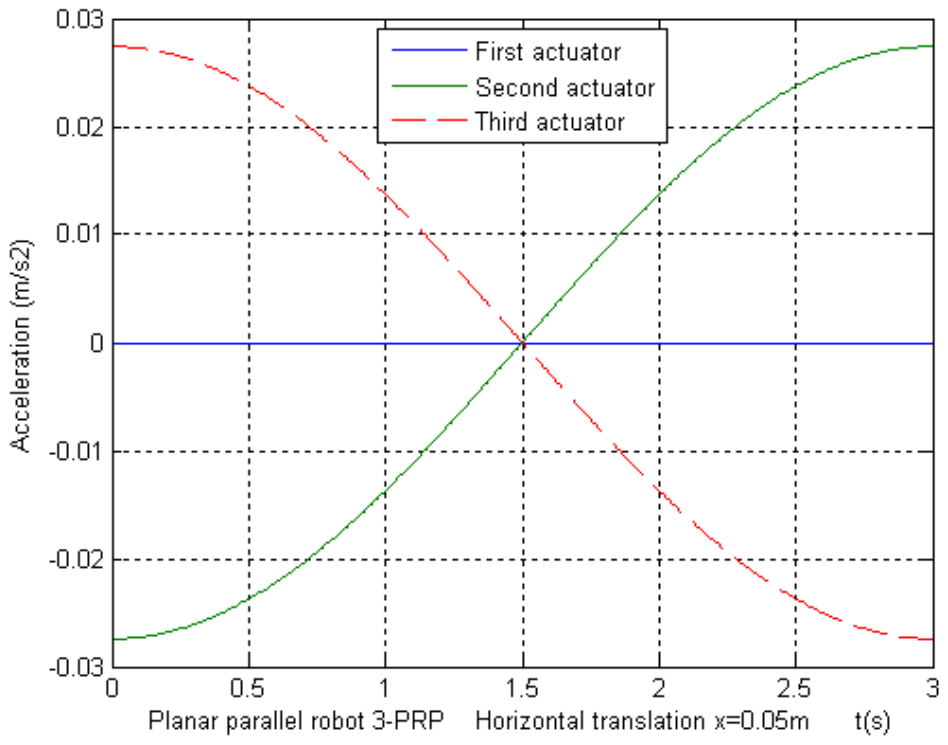

Fig. 8 Accelerations $\gamma_{10}^{A}$, $\gamma_{10}^{B}$, $\gamma_{10}^{C}$

As can be seen from Figs. 3, 4, 5, it is remarked that during the rotation motion of the platform all displacements, velocities and accelerations of three actuators are identically distributed.

In a second example, the supposed motion of the platform is a *translation* on horizontal $x_0$ axis (Figs. 6, 7, 8).

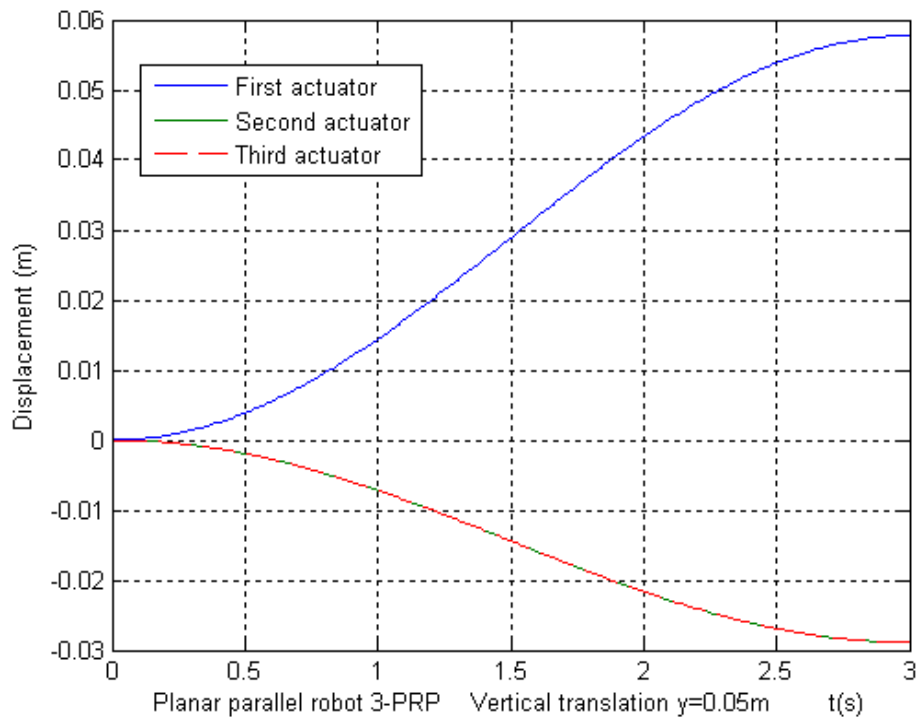

Fig. 9 Displacements $\lambda_{10}^{A}$, $\lambda_{10}^{B}$, $\lambda_{10}^{C}$

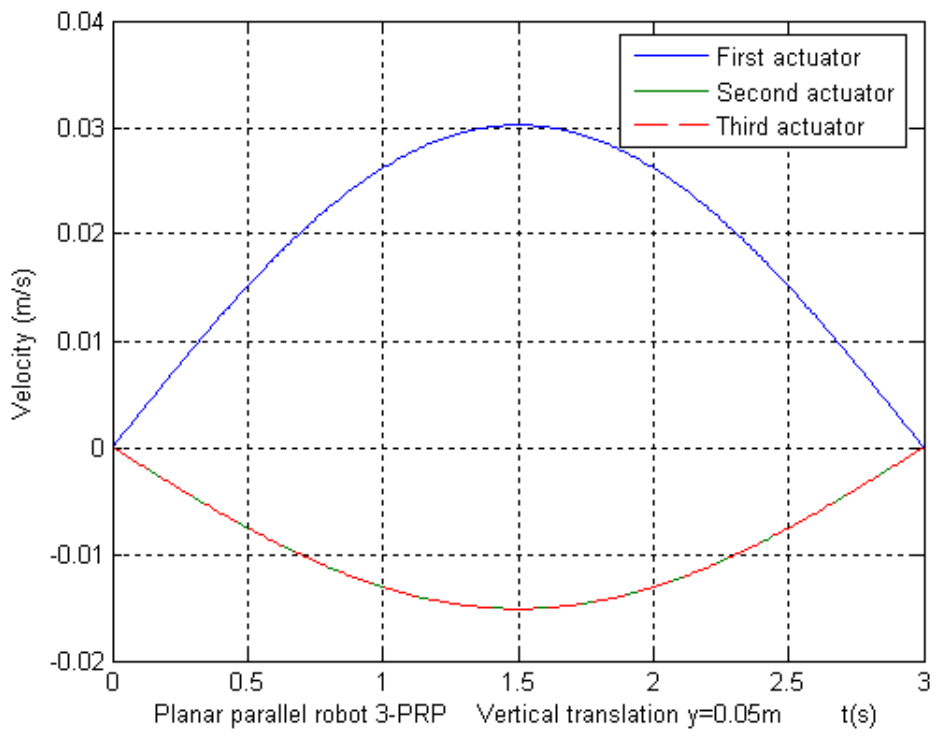

Fig. 10 Velocities $v_{10}^{A}$, $v_{10}^{B}$, $v_{10}^{C}$

Concerning the comparison in the case when the centre $G$ moves along a *rectilinear trajectory* along $y_0$ axis without any rotation of the platform, we remark that the distribution of displacement, velocity and acceleration calculated by the program and depicted in Figs. 9, 10, 11 is the same, at any instant, for two of three actuators.

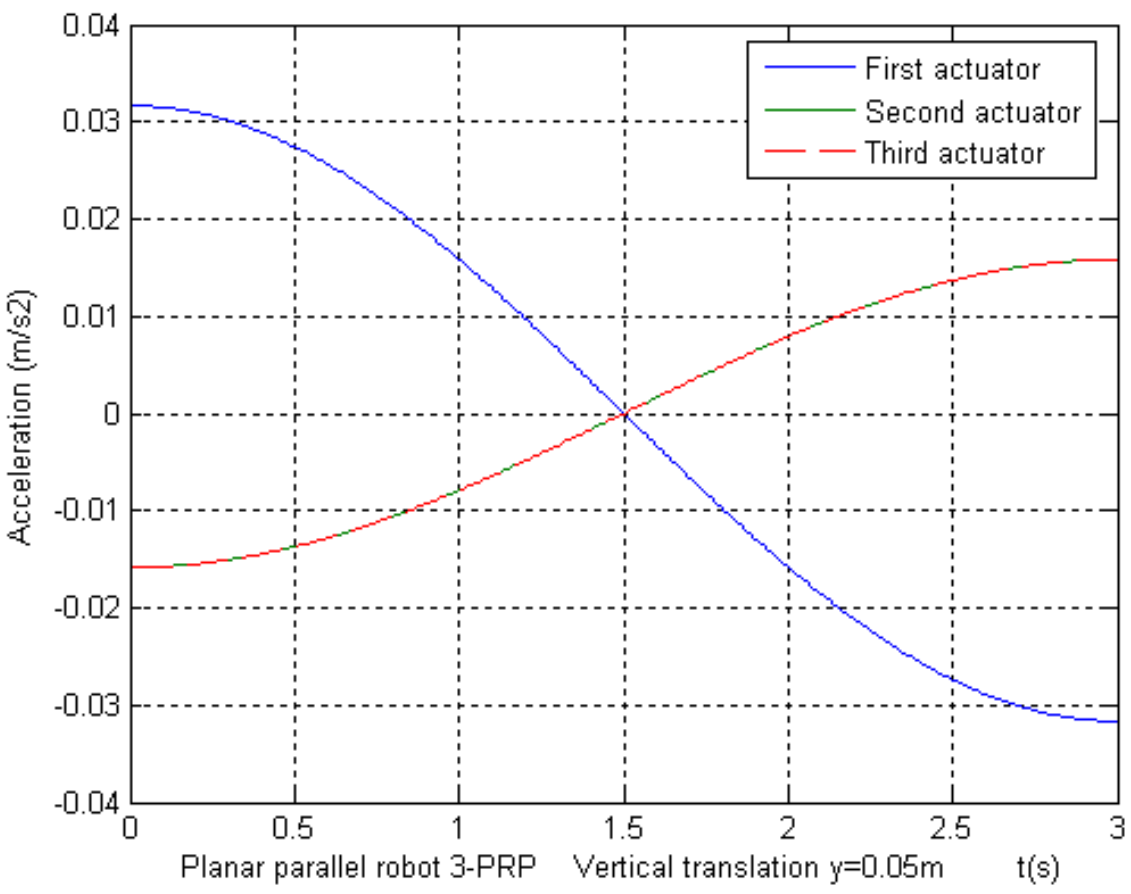

Fig. 11 Accelerations $\gamma_{10}^{A}$, $\gamma_{10}^{B}$, $\gamma_{10}^{C}$

The simulation through the MATLAB program certify that one of major advantages of the current matrix recursive approach it is the well structured way to formulate a kinematical model, which leads to a computational efficiency. The proposed method can be applied to various types of complex robots, when the number of components of the mechanism is increased.

## 3 Conclusions

Within the inverse kinematics analysis some exact relations that give in real-time the position, velocity and acceleration of each element of the parallel robot have been established in present paper. The method described above is quite available in forward and inverse mechanics of all serial or planar parallel mechanisms, the platform of which behaves in translation, rotation evolution or general 3-DOF motion.